\documentclass[10pt,letterpaper]{article} 
\usepackage[top=0.85in,left=2.75in,footskip=0.75in]{geometry}

\usepackage{amsmath,amssymb}
\usepackage{changepage}
\usepackage[utf8x]{inputenc}
\usepackage{cite}
\usepackage{nameref,hyperref}
\usepackage[right]{lineno}

\usepackage{microtype}
\DisableLigatures[f]{encoding = *, family = * }

\newcommand{\eref}[1]{Eq.~(\ref{#1})}

\newcommand{\fref}[1]{Fig.~\ref{#1}}
\newcommand{\Fref}[1]{Fig.~\ref{#1}}
\usepackage[table]{xcolor}
\usepackage{array}
\usepackage{caption}
\usepackage{multirow}
\usepackage{booktabs}
\usepackage{graphicx}
\usepackage{boldline,multirow}

\usepackage{soul}

\usepackage{pdfpages}
\usepackage{tcolorbox}
\usepackage{colortbl}
\usepackage[draft,inline,nomargin]{fixme} \fxsetup{theme=color}
\FXRegisterAuthor{jm}{jmm}{\color{purple}Josue}
\FXRegisterAuthor{cp}{acp}{\color{blue}CP}
\FXRegisterAuthor{cg}{cgg}{\color{red}CG}
\definecolor{C1-EN}{RGB}{0,132,170}
\definecolor{C1-FR}{RGB}{255,117,0}
\definecolor{C1-GE}{RGB}{134,0,179}
\definecolor{C1-IT}{RGB}{49,129,6}
\definecolor{C1-SP}{RGB}{162,0,0}

\newcolumntype{+}{!{\vrule width 2pt}}

\newlength\savedwidth

\raggedright
\usepackage[aboveskip=1pt,labelfont=bf,labelsep=period,justification=raggedright,singlelinecheck=off]{caption}

\makeatletter
\renewcommand{\@biblabel}[1]{\quad#1.}
\makeatother

\usepackage{lastpage,fancyhdr,graphicx}
\usepackage{epstopdf}
\fancyheadoffset[L]{2.25in}
\fancyfootoffset[L]{2.25in}
\begin{document}
\vspace*{0.2in}

\begin{flushleft}
{\Large
	\textbf\newline{Statistical analysis of word flow among five Indo-European languages} 
}
\newline
Josué Ely Molina  \ddag\textsuperscript{1,2}, 
Jorge Flores      \dag \textsuperscript{2}, 
Carlos Gershenson \textsuperscript{3,4,5},
Carlos Pineda     \ddag\textsuperscript{2,*}    

\bigskip
\newcommand{\unam}{Universidad Nacional Aut\'{o}noma de M\'{e}xico, Mexico City, 01000, Mexico}
\textbf{1} Facultad de Ciencias, \unam \\
\textbf{2} Instituto de F\'{\i}sica, \unam \\
\textbf{3} Instituto de Investigaciones en Matem\'{a}ticas Aplicadas y Sistemas, \unam \\
\textbf{4} Centro de Ciencias de la Complejidad, \unam \\
\textbf{5} Santa Fe Institute. 1399 Hyde Park Rd., Santa Fe, NM 87501, USA\\
\bigskip
\ddag These authors contributed equally to this work.
\dag Deceased \\
* carlospgmat03@gmail.com
\end{flushleft}
\section*{Abstract} 

A recent increase in data availability has allowed the possibility to perform different statistical linguistic studies. Here we use the Google Books Ngram dataset to analyze word flow among English, French, German, Italian, and Spanish. We study what we define as ``migrant words'', a type of loanwords that do not change their spelling. We quantify migrant words from one language to another for different decades, and notice that most migrant words can be aggregated in semantic fields and associated to historic events. We also study the statistical properties of accumulated migrant words and their rank dynamics. We propose a measure of \emph{use} of migrant words that could be used as a proxy of cultural influence. Our methodology is not exempt of caveats, but our results are encouraging to promote further studies in this direction.

\section*{Introduction} 


In recent years, the increase of data availability~\cite{Hilbert60} and the development of computational tools~\cite{doi:10.1098/rsta.2019.0061} has
benefited various statistical studies to understand certain characteristics of
the human population. For example, we are able to predict with a high confidence the growth rate of a
city~\cite{Batty:2012Cities,Murcio2015}, the number of people who have watched a movie~\cite{sinha2005blockbusters}, the user traffic on a web
page~\cite{BARABASI200069}, and even the way we use words in written language~\cite{Montemurro2001567,ZipfRnd2002}. The previous examples
are cases of  Zipf's law, formulated by George Zipf in the 1930s~\cite{Zipf,Petruszewycz1973Lhistoire-de-la,newman2005power,1367-2630-13-4-043004,Perc07122012,1367-2630-15-9-093033} upon
discovering that if  the words used in a text are ranked by their frequency of
appearance, where the lower ranks belong to the most frequent words,  then the
frequency  $f$  of any word and its rank  $k$ are related by a power law of the
form $f\sim1/k$.


Zipf's law has been mostly used to study the structures of language.
Nonetheless, not enough studies have been made to understand the historical and
cultural features that language provides. One way to begin such a study is by
noting that the languages themselves are mixed, since within the vocabulary of
a language, words from other languages are continuously added.


Currently in the Spanish language, there are loanwords from English 
that do not have a translation or that sometimes displace those that already
exist in Spanish.  For example, for native Spanish speakers in Mexico, it is common to
hear the word \textit{marketing} instead of its translation \textit{mercadeo}
when dealing with economic or business issues; also the word \textit{online}
has replaced \textit{en línea}, when referring to issues related to the
\textit{Internet}, a word officially adopted in Spanish.

This trend has not only affected Spanish,  but also other languages that are
being influenced by topics where English is the main and common language for
communication. However, in different periods of time, the flow of words came
from other languages. D’Amore~\cite{Damore_influencia_mutua} discusses with linguistic rigor the
flow of words between English and Spanish,  showing historical  and cultural
causes that allowed such a flow; in addition to mentioning the previous influence of
Arabic in Spanish and French in
English~\cite{gorlach2005dictionary,haspelmath2009loanwords,durkin2014borrowed}.

In this work, we use the Google Books N-gram dataset~\cite{ngramv} of the most
frequent words in books published in English, French, German, Italian, and
Spanish. With this dataset,  we develop an algorithm that identifies
the words of one language and that are being used exactly with the same 
spelling by others. Once these words
have been classified, we construct two models to quantify the influence that
one language has had on another during the 20th century. In the first model, we
count the number of new words that a language received from another. In
the second model, we develop the concept of the use of one language in
another, from quantifying the relative frequency of  the words of a language
that are being used in another language. In both models, we identify historical, social,
and cultural causes that are responsible for the flow of words.

Next, we use the concept of rank diversity~\cite{iplosone},  that shows the number of words
occupying a certain rank across the time. This study shows that  regardless of
the original or receiving language,  the lower ranks are always occupied by
fewer words, and as the rank increases, the diversity curve also increases following a sigmoid curve. 

Our work is of a statistical nature, and as such, has its limitations. However, we consider that statistical studies like the one we present can be complementary to detailed linguistic studies of loanwords. Certainly, we do not attempt to replace such studies, but to add insights and suggest further avenues of research. We do not need to sacrifice precision or large amounts of data~\cite{Harford2014} when we can have both.

In the next section, we present our methodology. Then, we show results and discuss for new migrant words per decade from the 1900s to the 2000s. We also analyze accumulated migrant words and their use. Afterwards, we study the rank dynamics of migrant words. Finally, we measure the robustness of our results by removing migrant words and comparing the resulting sets with the original ones. A discussion closes the paper.

\section*{Methodology} 

We used the Google Books $N$-gram dataset~\cite{ngramv}.
This dataset contains the usage frequency, for each year and language, of
the most used ``$N$-grams'' in Google Books. 
$N$-grams are the words or set of words that make up the text of a
book, where the number $N$ indicates the number of words that make up the gram,
being a 1-gram an individual word, a 2-gram a pair of words,
a 3-gram a sequence of three words, and so on.
We removed certain words that did not contributed to the analysis: articles,
pronouns, propositions, and conjunctions (all  of which are functional words),
since these serve to give a structure to the message. Then, we consider content
words such as nouns, main verbs, adjectives, and adverbs. 
From this dataset, and after cleaning the data, the lists of the five thousand
most used 1-grams each year
between 1740 and 2009 were extracted for the English, French, German, Italian,
and Spanish languages.
We are performing this cut as all the lists of the five languages (between 1740
and 2009), have at least this amount of 1-grams.
In each list, the words are ranked according to their
frequency of appearance, where the most frequent words have the lowest ranks.

To determine the presence of one language in another, an algorithm was
developed to find the words that are common between at least two languages,
these must have exactly the same spelling. These words were defined
as \textit{migrant words}, which are a particular case of loanwords (with identical spelling).

A migrant word is associated with a \textit{source language} and a
\textit{receiving language}, where the source language is the one where the
word appeared for the first time within the most used words,
while the receiving language is the one where the word is also present, being a
different set from the source language. 
To determine the source language, we established that this will be the language
where the word appeared for the first time within the five thousand most used
words. If a migrant word has appeared in the same year in two or
more languages, the source is the one where the word has the lowest rank.

The previous criterium for searching words with the same spelling and later
associating them with a source language is not perfect. There are some cases
that our method did not detect and were established as mistakes. One of the
most common errors was finding words with the same writing, but with different
meanings (polysemy). For example,  \textit{mayor} in English refers to the representative
of the government in a locality, while in Spanish, \textit{mayor} is an
adjective to indicate that something is greater, bigger, or older.  Another recurring
error was not distinguishing words with the same meaning but with slightly different spellings. For example, the word \textit{imagine} is written \textit{imaginer} in
French and \textit{imaginar} in Spanish.  Finally, in some cases, the authentic
source language is some other language for which there is no information in the
dataset, for example the word \textit{natural} comes from Greek, but there is no data
from Greek in the Google Books $N$-grams dataset. Consequently, our algorithm associated this word with English as its source language.

The above errors were detected by individually analyzing each of the migrant
words and their corresponding source and receiving languages. One way to have
cleaner data is by consulting an expert in each language, who reviews the words
and decides which ones were classified properly. However, this is not practical
since if there were more languages in the database, it would be necessary to
consult an expert for each language. Notwithstanding of this requirement to
regulate errors, we established a method to determine the importance (weight)
of these errors in the results, that will be shown in the following sections. 
\section*{New words} 
 
The purpose of this work is to establish the influence that one language has on
another. A first method to quantify such influence is by counting the
\textit{new migrant words} ($NMW$). These are words that appear for the first time in a
receiving language and that come from a unique source language.

\begin{figure} 
\begin{adjustwidth}{-1.2in}{0in}
\includegraphics[scale=.35]{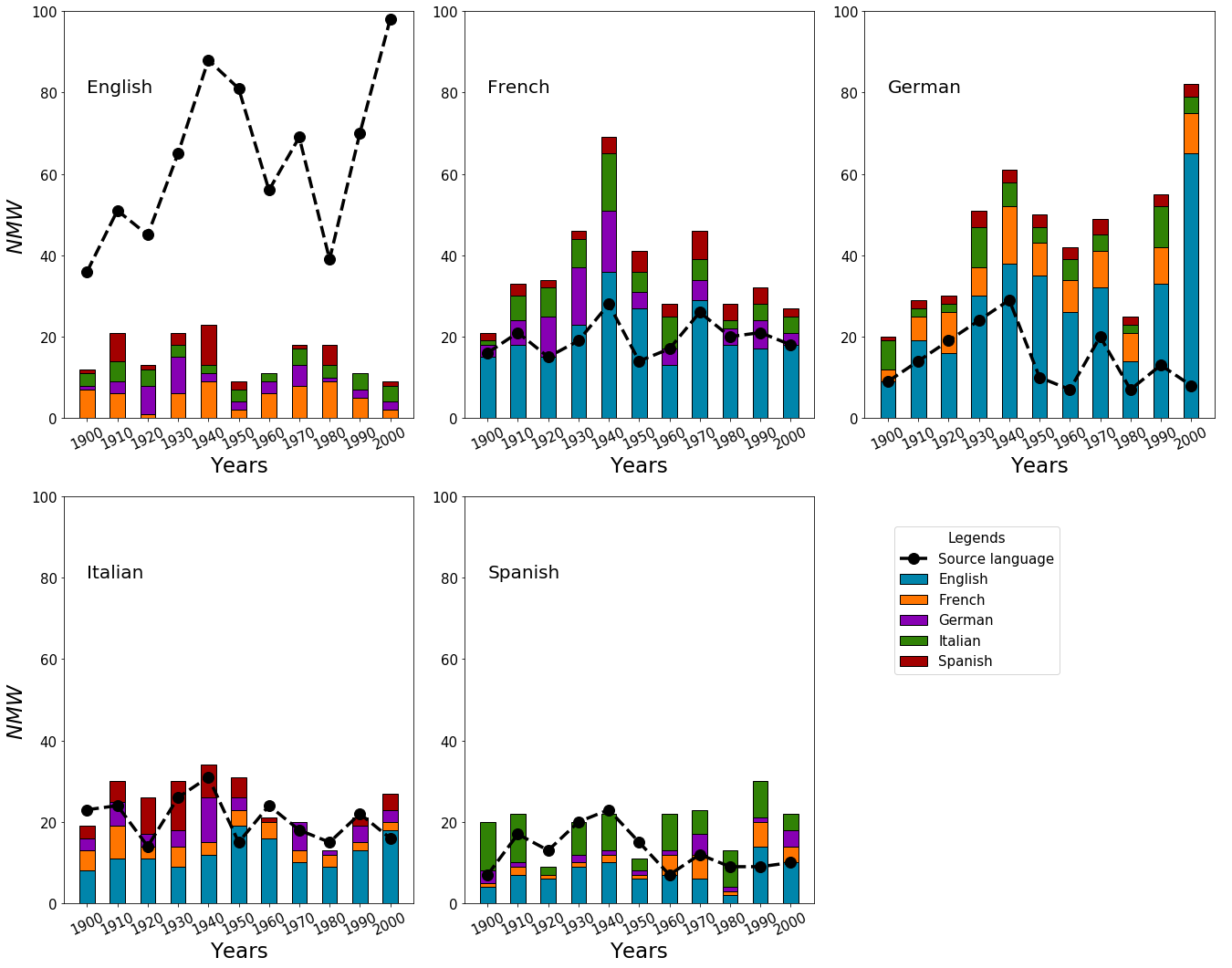}
\caption{{\bf New migrant words, per language and per decade.} 
$NMW$ are consider for each language. Each panel contains one language. 
The dotted line displays the number of $NMW_{out}$ that originate in the 
corresponding language, and the bars the $NMW_{in}$ coming to that language, 
separated by the origin of the different $NMW$. }
\label{fig.NMW_A}
\end{adjustwidth}
\end{figure}

We study the flow of $NMW$, per decade, in two ways. First, we
count the number of $NMW_{out}$ that a fixed language exports as a source
language. Second, counting, for a fixed language, the number
of new migrant words $NMW_{in}$. In this second way, we can study from 
which language are the $NMW_{in}$ coming.  The results are presented in 
\Fref{fig.NMW_A} for each decade of the 20th century. 

%

From this figure, we can see that the English language has
migrated on average three times more words than it has received, where the
greatest influence of English occurred in the 1940s and 2000s.
Consequently, the largest proportion of migrant words in the other languages come
from English. It is worth noticing that French, German, Italian, and Spanish exported more words
during the 1940s, but their export rate has remained roughly stable, with minimums for English in the 1900s and 1980s, French in the 1950s, German in the 1960s and 1980s, Italian in the 1920s, 1950s, and 1950s, and Spanish in the 1900s and 1960s.

The major influencer of English has varied across time, including German, Spanish, and French. Apart from English, French has received more influence from German and Italian, German from French and Italian, Italian from Spanish, and Spanish from Italian. Thus, it could be said that the second most influential language among the five studied has been Italian.

Analyzing the lists  of 
migrant words, we realized that
these can be grouped into semantic fields. According to~\cite{semantic_oxford},
a semantic field is a set of words that are related based on their meaning.
Table~\ref{tab.new_words} shows the words grouped by semantic fields, as well as
the pairs of source language and receiving language involved. We note that the
year of appearance (or in the years around) of the migrant words, a historical
or cultural event occurred that is related to the semantic field. For example,
between the 1930s and 1940s, words historically related to the Second World War
migrated between all languages; while since the 1990s, migrant words refer
to the fields of technology and globalization.  

\begin{table}[htb] 
\centering
\resizebox{\textwidth}{!}{
\begin{tabular}{@{}cV{4.5}cV{4.5}cV{4.5}c@{}}
\textbf{Semantic Field} & \textbf{New migrant words}
      & \textbf{Source language} & \textbf{Receiving language (decade)} \\
\hlineB{4.5}
\multirow{2}{*}{World War I} 
   & \begin{tabular}[c]{@{}c@{}} \\ allemagne,  austro,\\  russie, versailles.\end{tabular} 
   & FR    & \begin{tabular}[c]{@{}c@{}}\\IT (1910, 1920), GE (1920).\end{tabular} \\
 & kaiser, reich. & GE   & FR (1910). \\
\hlineB{4.5}
	\multirow{5}{*}{World War II}                                                                    & \begin{tabular}[c]{@{}c@{}}catastrophe, churchill,\\ nazis, stalin, territories.\end{tabular}                                                           & EN                       & \begin{tabular}[c]{@{}c@{}}FR (1940), \\ IT (1940).\end{tabular}                                                     \\
	& \begin{tabular}[c]{@{}c@{}}berchtold, bestimmungen,\\  hitler, kaiser, lenin, \\ minister, regierung.\end{tabular}                                      & GE                       & \begin{tabular}[c]{@{}c@{}}EN (1930), FR (1940),\\ IT (1930, 1940),\\ SP (1930).\end{tabular}                        \\
	& \begin{tabular}[c]{@{}c@{}}duce, mussolini, \\ regime.\end{tabular}                                                                             & IT                       & \begin{tabular}[c]{@{}c@{}}EN (1930), FR (1930),\\ GE (1930, 1940),\\ SP (1930).\end{tabular}                        \\
	
	\hlineB{4.5}
	Aftermath of WWII                                                                                & onu, urss, vietnam                                                                                                                                      & FR                       & SP (1960, 1990).                                                                                                     \\
	\hlineB{4.5}
	\multirow{2}{*}{\begin{tabular}[c]{@{}c@{}}Historic figures\\ in arts, science\\ and philosophy \end{tabular}} & poincare.                                                                                                                                               & FR                       & IT (1920).                                                                                                           \\
	& \begin{tabular}[c]{@{}c@{}}bach, beethoven, engels, \\ freud, hegel, heidegger, \\ marx, mozart, nietzsche.\end{tabular}                                & GE                       & \begin{tabular}[c]{@{}c@{}}EN (1930), FR (1900-1980),\\ IT (1900, 1940, 1970),\\ SP (1930, 1970, 1980).\end{tabular} \\
	\hlineB{4.5}
	\multirow{2}{*}{\begin{tabular}[c]{@{}c@{}}Ideologies and \\ political terms\end{tabular}}       & \begin{tabular}[c]{@{}c@{}}burgueoise, diplomatie, \\ empire, politique.\end{tabular}                                                                   & FR                       & GE (1910-1930).                                                                                                      \\
	& \begin{tabular}[c]{@{}c@{}}capitalista, comunista, fascismo, \\ marxismo, socialista, terrorismo.\end{tabular}                                          & IT                       & SP (1910, 1930, 1960, 1980).                                                                                         \\

	\hlineB{4.5}
	Economy                                                                                          & \begin{tabular}[c]{@{}c@{}}depression, dollar, economic, \\ economy, financial, investment, \\ market, marketing, value.\end{tabular}                   & \multirow{4}{*}{EN}      & \begin{tabular}[c]{@{}c@{}}FR (1950), \\ GE (1930, 1990, 2000),\\ SP (1980-2000).\end{tabular}                       \\
	
	Technology                                                                                       & \begin{tabular}[c]{@{}c@{}}digital, internet, mail, \\ online, software.\end{tabular}                                                                   &                          & \begin{tabular}[c]{@{}c@{}}FR (1990), GE (1990),\\ IT (1990, 2000), \\ SP (1990, 2000).\end{tabular}                 \\
	Globalization                                                                                    & \begin{tabular}[c]{@{}c@{}}business, customer, management, \\ market, marketing, standars.\end{tabular}                                                 &                          & \begin{tabular}[c]{@{}c@{}}FR (2000), GE (1990, 2000),\\ IT (2000),  SP ( 2000).\end{tabular}                        \\
	\begin{tabular}[c]{@{}c@{}}Presidents of the\\ United States of\\p America\end{tabular}           & \begin{tabular}[c]{@{}c@{}}roosevelt, truman, kennedy,\\ johnson, nixon, reagan,\\ bush, clinton.\end{tabular}                                          &                          & \begin{tabular}[c]{@{}c@{}}FR (1960, 1970), GE (1930,1940),\\ IT (1940),  SP (1940-1990).\end{tabular}               \\
	\hlineB{4.5}
	Medicine                                                                                         & \begin{tabular}[c]{@{}c@{}}anemia, anestesia, colesterina,\\ endovenosa, gastrica, lepra\\ metabolismo, sintomatologia, virus,\\ vitamina.\end{tabular} & \multirow{2}{*}{SP}      & \begin{tabular}[c]{@{}c@{}}EN (1940), GE (1990,1940),\\ IT (1920, 1930),\end{tabular}                                \\
	\begin{tabular}[c]{@{}c@{}}Latinoamerican countries \\ and cities\end{tabular}                   & 
	\begin{tabular}[c]{@{}c@{}}argentina, aires, buenos,\\ chile, panama.\end{tabular}                                                                      &                          & \begin{tabular}[c]{@{}c@{}}EN (1910, 1940), FR (1910), \\ IT(1900).\end{tabular}                                    
\end{tabular}%
}
\caption{\textbf{New migrant words.} 
Examples of new migrant words for all pairs of languages, grouped together by semantic field.
A notorious influence of historic events on word migration is observed. 
We use the following abbreviations: EN for English, FR for French, GE for German, IT for
Italian and SP for Spanish. 
}
\label{tab.new_words}
\end{table} 

These kind of groupings allow us to say in which areas the languages are most
influential and the reason for the migration of words. The English language has
migrated words to others because of technological development and
globalization in the last thirty years. French, Italian and German
were influential after the war events of the 20th century, in addition to the academic
influence of Germany seen through surnames of historic figures. Finally, Spanish was influential after economic crises in Latin American
countries~\cite{crisis_chile}. The fact that locations from one country become frequent words in another language suggests that some people speaking the latter are interested in the former. Similarly, influence can be seen e.g. as USA presidents become commonly used words (in the top 5000) in other languages. This suggests that migrant words could be used as a proxy measure of cultural influence (see next Section).

\begin{figure}[!h]
	\begin{adjustwidth}{-2cm}{1cm}
		\centering
		\includegraphics{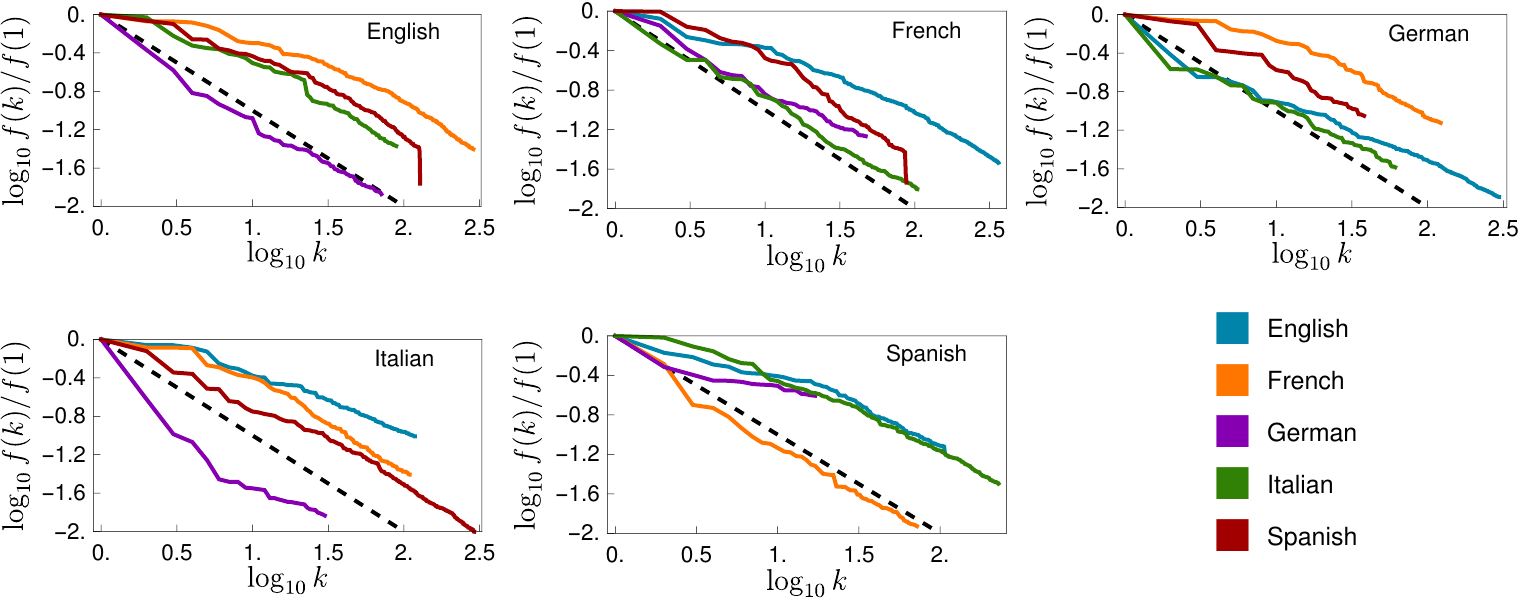}
		\caption{{\bf Zipf's law of the accumulated migrant words,
grouped by receiving language.} We display a frequency-rank 
plot for all language pairs, for the migrant words during the year 2000.
Indeed, after a transient, Zipf's law is observed (a dashed line with slope $-1$ is
provided for comparison). }
		\label{fig.ZL_receiving}
	\end{adjustwidth}
\end{figure}

Another interesting feature that we observed is that migrant words also 
fulfill Zipf's law~\cite{Zipf}. In Fig~\ref{fig.ZL_receiving} we present all language pairs, 
grouped by receiving language and we observe, 
within statistical fluctuations, an asymptotic power-law decay with an exponent
close to one. 

\section*{Accumulated words} 

The previous results show that words travel from one language to another in
groups belonging to a common semantic field. Nevertheless, we still cannot
associate them with a number that quantifies how much influence has one
language on another.

To obtain such a number, we will focus on migrant words in the years after the
first year they migrated, observing how their frequency varies over time. For
example, a migrant word will begin to be influential if its frequency increases
over time. Since we are dealing with groups of words, we define as accumulated
migrant words, those words with source language $A$ that already appeared in a
receiving language $B$, and for a given year they do so again.

Consider the words that up to a given year $t$ have migrated from language $A$ to $B$. Each of these
words will have a ranking $f(j)$ where $j$ is the ranking of the word, within the aforementioned
list of words. 
We now
add the frequencies of the accumulated
migrant words of $A$ to $B$ at a year $t$ and normalize the this quantity by
dividing it by the frequency of the first five thousand words that make up the list:
\begin{equation}
\label{ec.fuso}
\underset{ \text{\tiny A} \to  \text{\tiny B} }{U}(t) = \frac{\sum_{j}
f(j)}{\sum_{k=1}^{5000} f(k)}.
\end{equation}
We define this new value as the \textit{use} of $A$ in $B$, and interpret is value
as a measure of influence. It will then be said that the influence of $A$ has increased on
$B$, if in an interval of time $\Delta t$ the use of $A$ on $B$, $\underset{ \text{\tiny A} \to  \text{\tiny B} }{U}$, increases.

\begin{figure}[!h]
	\begin{adjustwidth}{-2cm}{1cm}
		\centering
		\includegraphics{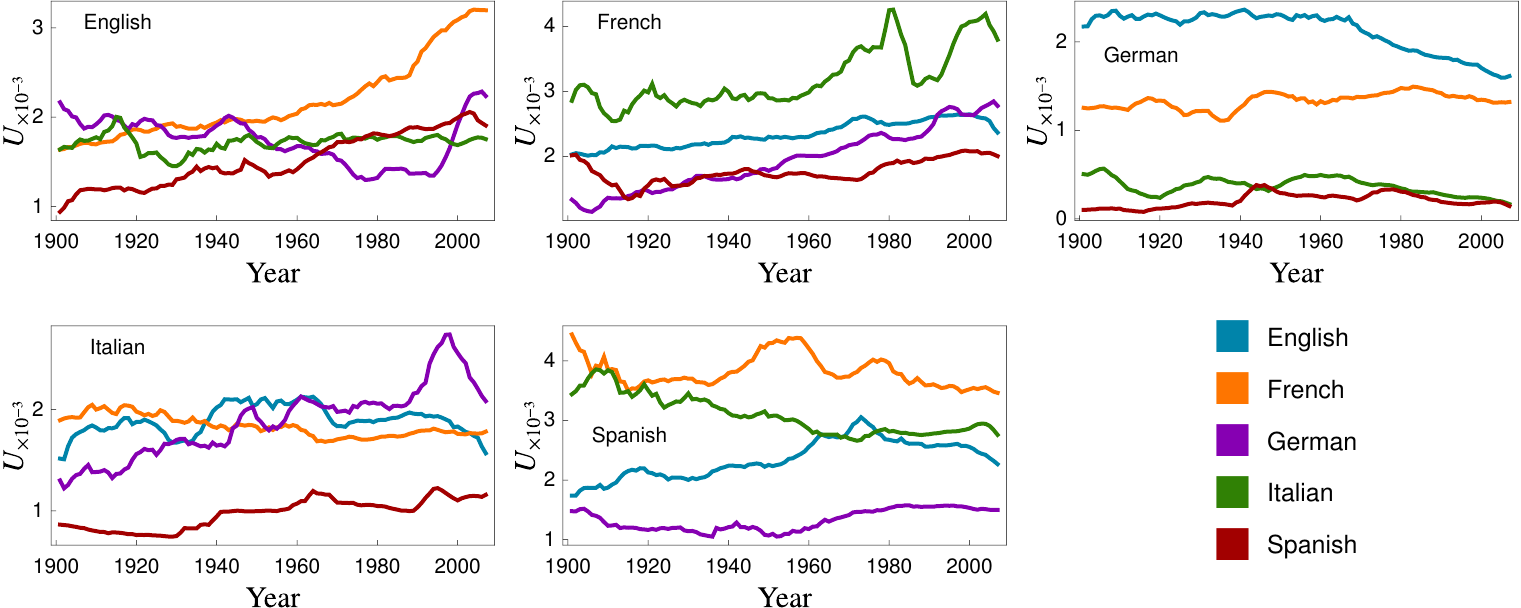}
		\caption{{\bf The use $U$ among languages.} 
We plot the use, as defined in \eref{ec.fuso} for all language pairs. 
Results are discussed in the
main text. }
		\label{fig.UT_art}
	\end{adjustwidth}
\end{figure}

We obtained the accumulated migrant words for all possible combinations of
source and receiving languages from 1740 to 2009.  Afterwards, we calculated
the use, \eref{ec.fuso},  for each pair of languages between 1900 and 2009, so
as to have a time period (1740-1899) to build a large enough dataset to have
meaningful migrant words. The results are presented in \fref{fig.UT_art}, 
grouped by source language.

\subsubsection*{English} 

The use of English in French and Spanish has increased steadily in the last century
whereas in Italian, it has maintained a constant level. 
An increase of its use in German occurred after 1990, just after the fall of the Berlin Wall. We associate the cause of these increases
with the emergence of the United States as a world power, after the end of the
World War II, the propagation of its economic model, as well as the development
of science and technology. The accumulated migrant words that are present in all receiving
languages are \textit{capital}, \textit{dollar}, \textit{invesment},
\textit{relations}, \textit{market}, \textit{company}, \textit{development},
\textit{financial},  \textit{institutions}, \textit{internet}, \textit{windows}
and \textit{software}. Those can again be associated with semantic fields, such
as economics, technology, and globalization.

\subsubsection*{French} 

The increase of French influence in the other languages occurred in
English between 1920 and 1970, in German between 1900 and 2009, and in Spanish
between 1970 and 1995. In these years, words that increased their frequency
are from the semantic fields of religion such as \textit{dieu} ,
\textit{évêque}, \textit{dime}, \textit{religion}, \textit{saint} and
\textit{église} ; while \textit{reine}, \textit{forteresse},
\textit{napoleon}, \textit{guerre}, \textit{imperiale}, \textit{bastille},
\textit{royal} and \textit{bourgeois} have their roots in French history.

In Italian, between 1950 and 1970, in addition to the above
words, \textit{raisins}, \textit{vin}, \textit{vignoble} and \textit{recolte} were
found, the common meaning of which is the wine industry, a common industry in
France and Italy.

\subsubsection*{German} 

Spanish and French,  between 1930 and 1945, were the languages where German had the greatest
increase among all receiving languages. 
In both, the words that are present are the surnames of German-speaking
influential figures, such as \textit{Marx},
\textit{Freud}, \textit{Heidegger}, \textit{Nietzsche}, \textit{Hegel},
\textit{Engels} and \textit{Mozart}.

In English and Italian, the biggest change was between 1960 and 2009, where the
use of German decreased. In this period, some words that lost influence, in the
sense that their frequency decreased. 
Among these words are 
\textit{Berlin}, \textit{Marx},
\textit{Hitler}, \textit{Lenin}, \textit{testen}  and \textit{reich}.
\subsubsection*{Italian} 

The influence of Italian came mainly from two semantic fields,  WWII with
\textit{Mussolini}, \textit{fascismo}, \textit{battaglia}, \textit{regime},
\textit{sociale}, and \textit{liberale}; and religion with \textit{santo},
\textit{suora} and \textit{cattedrale}. These semantic fields are
responsible for the increase in English between 1930 and 1940, in German
between 1950 and 1995, and in Spanish between 1930 and 1960.
A significant maximum of the influence of Italian in German is observed close to 1990. 
The words that produce such peak include 
\textit{usa},  \textit{tv}, and \textit{Rome}, which expose the limitation of our method. 
We think that the fact that 1990 FIFA World Cup final was played in Rome, where Germany won, might explain such behavior in the usage. 
%

In French, Italian has the lowest influence, where the aforementioned words
began to be less frequent between 1940 and 1960.
\subsubsection*{Spanish} 

The influence of Spanish in English between 1920 and 1970, was due to
historical and cultural facts. Names of Latin American  countries such as
\textit{Mexico}, \textit{Panama}, \textit{Chile}, \textit{Cuba}, \textit{Peru},
\textit{Colombia}, \textit{Argentina} and its capital \textit{Buenos}
\textit{Aires};  and states that were previously Spanish colonies, and maintain Spanish names, such as \textit{California} and
\textit{Florida}, are an important part of words that appeared in 
other languages. 

In German after WWII, and in French between 1930 and 1955, the main words
involved in that increase are, \textit{terapia}, \textit{anemia},
\textit{lepra}, \textit{tumor}, \textit{syphilis}, \textit{virus}, and
\textit{renal}, related to the medical semantic field.

\section*{Rank diversity} 

In the previous sections, we quantified the influence of a language on another. However, 
one can wonder about how the migrant words change in time. Are the most important words
the same, or do they change? In fact, 
since the accumulated migrant words are organized by year, and at the same time
in each year the words are ordered in ascending order in rank, then over time,
the same rank can be occupied by different words. One way to quantify
this change is through rank diversity $d(k)$~\cite{iplosone}. This quantity is
defined as the number 
of different elements that occupied rank  $k$ within the same dataset, divided
by the number of time slots considered.
Rank diversity has been used in datasets of the most used words in six
Indo-European languages~\cite{iplosone,10.3389/fphy.2018.00045,Cocho2019}, in sports and game classifications \cite{Morales_epj}, 
and in many other datasets~\cite{Iniguez2022}. 
Although in previous studies of rank diversity of languages and the current one the criteria for establishing rankings are different, in both there
is a common result: the lowest ranks are always occupied by fewer
elements, thereby as the rank increases, the number of different elements that occupied it
also does.

\begin{figure} 
\centering
\includegraphics[scale=1]{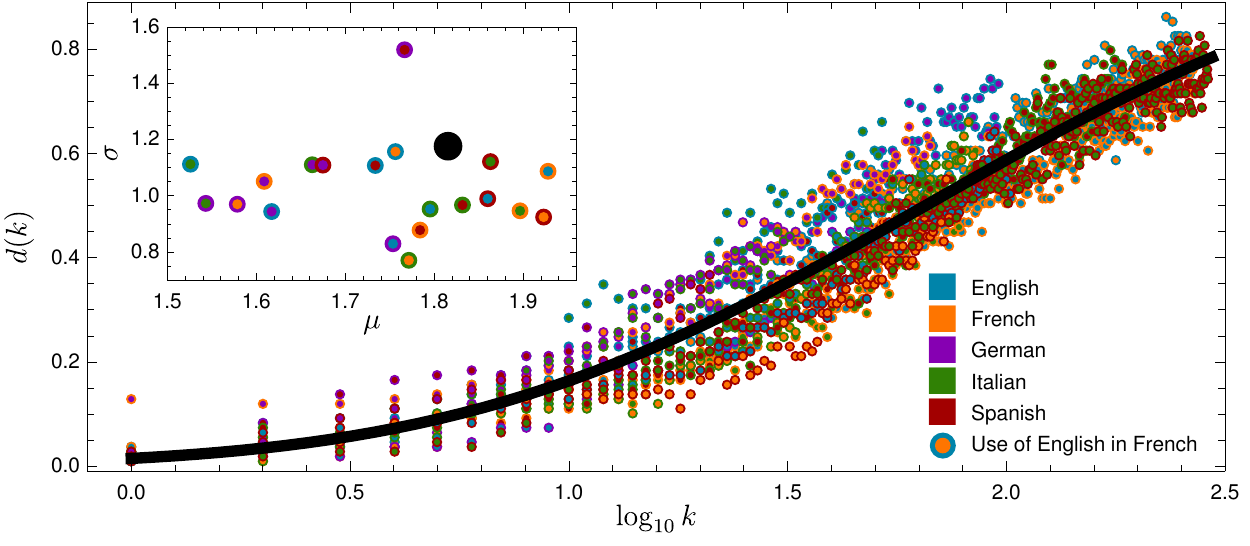}
\caption{{\bf Rank diversity of accumulated migrant words among languages.} 
Diversity for all pairs of languages. Each pair is well fitted by the 
sigmoid proposed in \eref{eq:sigmoid}, with fitting parameters $\mu$ and $\sigma$
reported in the inset. As a reference, we show a global sigmoid (in black)
obtained by fitting all data points. Its fitting parameters are also shown in the
inset as a black dot. 
%
%
%
}
\label{fig.DR_art}
\end{figure} 
After calculating the rank diversity (considering all years) for each source and receiving
language pair, the diversity values resemble a sigmoid curve, as can be
seen in Fig~\ref{fig.DR_art}. This can be fitted with a curve that is
cumulative of a Gaussian centered at $\mu$ and with 
deviation $\sigma$, i.e.  
\begin{equation}
\Phi_{\mu,\sigma}(\log_{10} k)=\frac{1}{\sigma\sqrt{2\pi}}
   \int_{-\infty}^{\log_{10}k}\exp\left(-\frac{(y-\mu)^2}{2\sigma^2}\right) {\rm d} y,
   \label{eq:sigmoid}
\end{equation}
The parameters $\mu$ and $\sigma$
are obtained with a linear regression. 
It is observed that the behavior of
diversity increases as the rank also increases, regardless of whether the
corpus has few or many ranks ($14$ in German-Spanish, $290$ in Spanish-Italian,
etc). With this, it can be concluded that, the migrant accumulated words in the
middle and high ranks are the ones that tend to change their position the most
within a ranking over time.

These observations suggest that only (relatively) few migrant words are used frequently, during long periods of time, while most migrant words are used not so (relatively) frequently, and their usage varies (relatively) more with time. 


\section*{Robustness} 
The method we used to build the set of migrant words relied on words having {\it
exactly} the same spelling when going from one language to another. We know 
that that is not always the case. Some words change their spelling. For example
the word {\it parquear} in Spanish comes from to the verb {\it to park} in
English.

To check the stability of the results presented and the importance of omitting
certain words, we proceeded as follows:  Take the original set (the one
used in the previous section) of the accumulated words of a pair of source
language and receiving language. From this set, eliminate a certain group
of words, in order to obtain a reduced set; in both,  equation~\ref{ec.fuso} is
used to obtain the modified use between the years 1900 and 2009.  The next
thing is to determine how similar the use of both sets are. We normalized the
values of both sets, after dividing them by the average value of each one; then
for each year $t$ we obtain the distance between each value of original use
$u_{t}$ and its corresponding value in reduced use $r_{t}$. The average of them
gets the \textit{average distance} $\left\langle D \right\rangle$, which will
be the one that quantifies the similarity of the results, indicating a greater
similarity if it is close to zero. This distance is defined as 
\begin{equation}
\left\langle D \right\rangle  = \frac{1}{N}\sum_{t=1}^{N} \left| u_{t} - \tilde u_{t} \right|  
\label{ec.Davg}
\end{equation}
where $u(t)$ is the original normalized usage and $\tilde u(t)$ is the reduced,
normalized usage.


Recalling that migrant words have frequency inversely proportional to the rank,
it is clear that some words are more important than others (see
\fref{fig.ZL_receiving}).  Thus, care must be taken when one removes a fixed
proportion of words, or a fixed frequency, as it can cause a big difference. 
One way to explore such aspect is to remove words from higher ranks
or lower ranks. 
With these ideas, in each source language
and receiving language pair, we carry out two types of elimination, in the
first we begin to eliminate the words with the lowest ranks gradually 
increasing the proportion of words removed $R_{p}$ (from 1$\%$ to a
99$\%$); in the opposite way, for the second case,  we begin by
eliminating those words with highest ranks. In both cases,
each time the eliminated portion was increased, the average difference was
calculated to observe the similarity.

\begin{figure}[!h]
	\begin{adjustwidth}{-2.5cm}{1cm}
		\centering
		\includegraphics{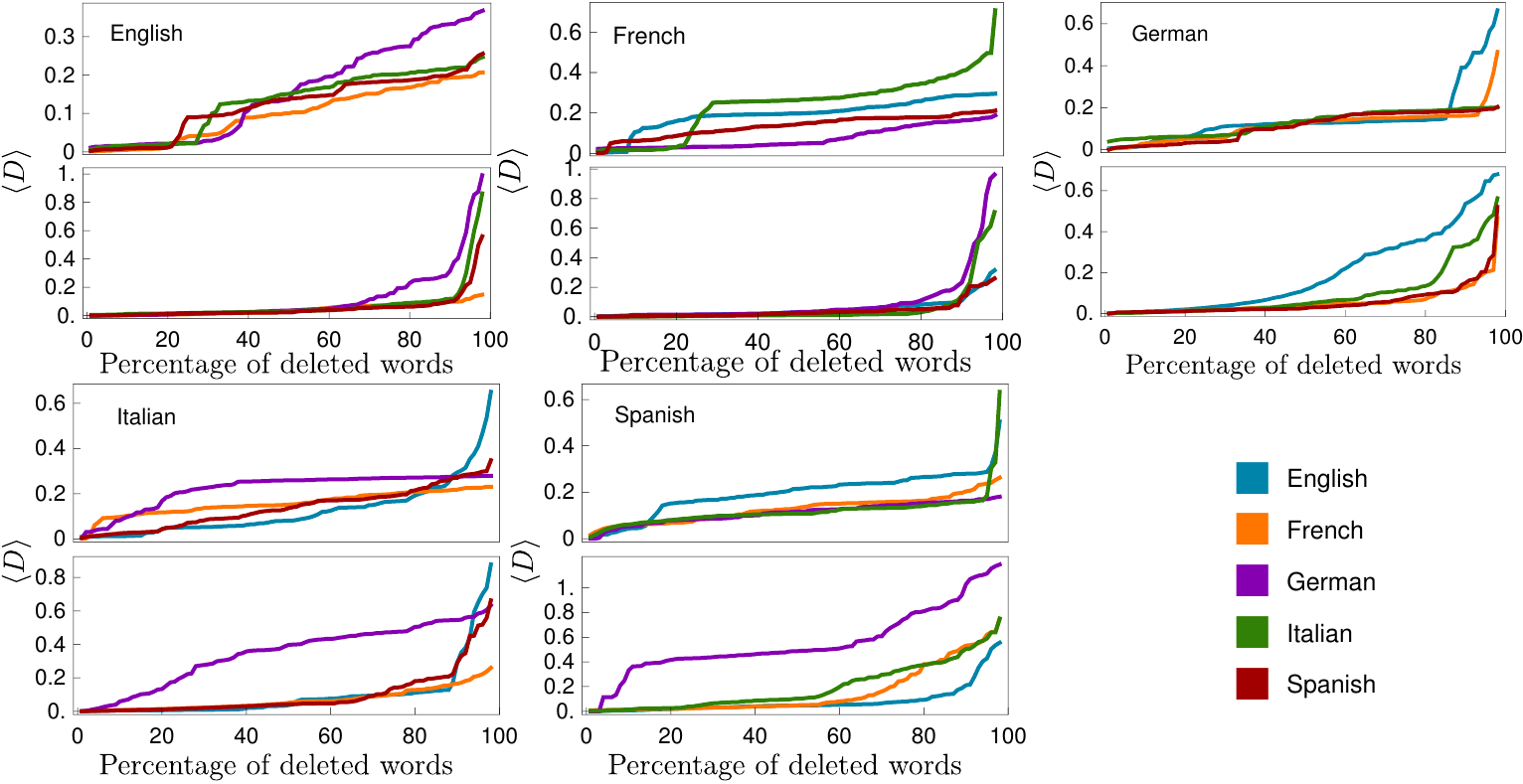}
		\caption{{Similarity of use when some words are eliminated.}
We study the similarity of the shape of usage \eref{ec.fuso}, as measured by
$\left\langle D \right\rangle $ (see \eref{ec.Davg}), when some words are eliminated.
The upper plots correspond to the case in which the lower ranked words (more frequent) are being
eliminated first, whereas in the lower plots we start eliminating the higher ranked
words (less frequent). 
}
		\label{fig.RP}
	\end{adjustwidth}
\end{figure}

%

In Fig.~\ref{fig.RP} we can observe how much the shape of the curves for usage changes, 
with an increasing proportion of words eliminated. Clearly, when removing the 
lower ranking words, the deformation is greater. However, we see that in 
most cases,  removing the 60\% of the higher ranked words produces
a deformation with $\langle D \rangle < 10\%$ (exceptions being 
German influencing English, Italian influencing German and
Spanish influencing German and Italian). 
%
%
%
%
This result implies that care must be taken when doing this analysis with respect to 
words that have low ranks. Thus, using this automated approach
to yield quantitative statements should pay special attention to the most frequently used words. 

%


\section*{Discussion} 

We presented a method and analyzed how migrant words (loanwords with the same spelling) have spread across five Indoeuropean languages. This ``blind big data'' approach can offer some insights, so as to estimate influences in different temporal periods, and how historic events might have contributed to these fluxes. Nevertheless, our method has important caveats. As it is, it requires that languages use the same alphabet (although some automatic transliteration could be used to include more languages, such as Russian). However, languages also change the spelling of words as they migrate from one to the other. For example, in Spanish it is difficult to pronounce words beginning with `s' followed by a consonant, so an `e' is added (e.g., `especial', `espectacular', `estable'; although these come from Latin, not English). Indeed, some words might have their origins in a language not considered in a specific study, e.g. `sushi', although it might be useful to learn in which languages such words became popular first, even when it is uncertain whether the word migrated directly from the original to others, or via an intermediary (such as Nahuatl words that have spread globally through Spanish, e.g. `tomatl', `xocolatl'). Still, as more data becomes available, more languages could be included in statistical studies. Another limitation of our present work is that we focus on frequently used words, although the same methodology could be used for less frequent words as well. These caveats imply that we are not attempting to replace ``insightful small data'' studies, where experts focus on particular words and study how they are shared across languages. We believe that both types of studies are complementary and necessary.

Additionally from the linguistic aspects of these studies, they can be useful to study cultural influence as well. The fact that a name of a place or person from one place is used frequently in another implies relevance. Thus, migrant words can be used also as proxies of cultural influence.

More sophisticated statistical linguistic studies are becoming possible because of increasing data availability and computational processing power. Still, we must be aware of the limitations of these methods. They can offer useful insights, complementary to but not replacing other approaches in linguistics and culturomics~\cite{Michel14012011,doi:10.1073/pnas.2102061118}.

\section*{Supporting information} 
\section*{Acknowledgments} 

\nolinenumbers
We are grateful for stimulating conversations with Sergio Sanchez during the onset of this project.  Support by projects CONACyT 285754 and UNAM-PAPIIT (IG101421, IN107919, IV100120, IN105122) and from the PASPA program from UNAM-DGAPA is acknowledged. 
\bibliography{referencias} 
\end{document}